\documentclass[sn-nature]{sn-jnl}

\usepackage{graphicx}%
\usepackage{multirow}%
\usepackage{amsmath,amssymb,amsfonts}%
\usepackage{amsthm}%
\usepackage{mathrsfs}%
\usepackage[title]{appendix}%
\usepackage{xcolor}%
\usepackage{textcomp}%
\usepackage{manyfoot}%
\usepackage{booktabs}%
\usepackage{algorithm}%
\usepackage{algorithmicx}%
\usepackage{algpseudocode}%
\usepackage{listings}%
\usepackage{hyperref}
\usepackage{float}

\begin{document}

\title[Article Title]{Airfoil Diffusion: Denoising Diffusion Model For Conditional Airfoil Generation }

\author[1]{\fnm{Reid} \sur{Graves}}\email{rgraves@andrew.cmu.edu}

\author*[1]{\fnm{Amir} \sur{Barati Farimani}}\email{barati@cmu.edu}

\affil[1]{\orgdiv{Department of Mechanical Engineering}, \orgname{Carnegie Mellon University}, \orgaddress{\street{5000 Forbes Avenue}, \city{Pittsburgh}, \postcode{15213}, \state{PA}, \country{USA}}}

\abstract{The design of aerodynamic shapes, such as airfoils, has traditionally required significant computational resources and relied on predefined design parameters, which limit the potential for novel shape synthesis. In this work, we introduce a data-driven methodology for airfoil generation using a diffusion model. Trained on a dataset of preexisting airfoils, our model can generate an arbitrary number of new airfoils from random vectors, which can be conditioned on specific aerodynamic performance metrics such as lift ($C_l$) or drag ($C_d$). Our results demonstrate that the diffusion model effectively produces airfoil shapes with realistic aerodynamic properties, offering improvements in efficiency and flexibility. The model produces within the design space, and additionally generates new samples in sparse regions of the design space, with the potential to find new designs that exhibit favorable performance characteristics.
}


\maketitle
\vspace{0.5cm} 
\noindent\textbf{Article Highlights}
\begin{itemize}
    \item A novel denoising diffusion model generates airfoil shapes that could improve aerodynamic performance.
    \item The model can be steered towards generation of profiles with specific lift and drag characteristics.
    \item This approach expands potential design techniques in aerospace applications.
\end{itemize}

\section{Introduction}\label{sec1}

The design and discovery of aerodynamic shapes, particularly airfoils, are fundamental challenges in aerospace engineering. Traditionally, the process of designing airfoils has relied heavily on computationally intensive simulations to explore the vast design space \cite{review_optimization_methods, raymer2006aircraft}. This approach typically involves defining specific design parameters that can limit the ability to synthesize truly novel shapes \cite{Wang2021Airfoil}. As the demand for higher performance and more efficient aerodynamic structures increases, there is a growing need for advanced methodologies that can streamline the design process and facilitate the generation of innovative airfoil geometries \cite{Morris2008Wing}.

Recent advances in machine learning have opened new avenues for the design and optimization of complex systems \cite{Hermans2014Automated}. Data-driven approaches, such as generative models, have shown promise in various fields by learning from existing data to generate new high-quality samples\cite{jadhav2023stressd, jadhav2024generative}. In the context of airfoil design, these models can potentially revolutionize the way we approach shape generation by learning the underlying distribution of airfoil geometries and their aerodynamic properties \cite{Wang2021Airfoil, Chen2021Deep, Xie2022Soft}.

In this work, we propose a novel methodology for airfoil generation using a diffusion model. Using a pre-existing dataset of 2 dimensional airfoils, we train a diffusion model that synthesizes novel airfoil profiles.

The diffusion model, once trained, allows for the generation of new airfoil shapes from random vectors. This capability enables the synthesis of new airfoils that can be selected on the basis of desired Cl or Cd values. Our approach offers a data-driven framework that not only enhances the efficiency of the airfoil design process, but also has the potential to discover novel shapes with superior aerodynamic performance within the training dataset design space.

This paper is structured as follows. Section 2 provides an overview of related work in the field of generative models for aerodynamic design. Section 3 details the methodology, including the architecture of airfoil diffusion and the training process of diffusion models. Section 4 presents the results of our experiments that demonstrate the effectiveness of our approach. Finally, Section 5 concludes the paper and discusses potential future work.

\begin{figure}[htbp]
    \centering
    \includegraphics[width=\linewidth]{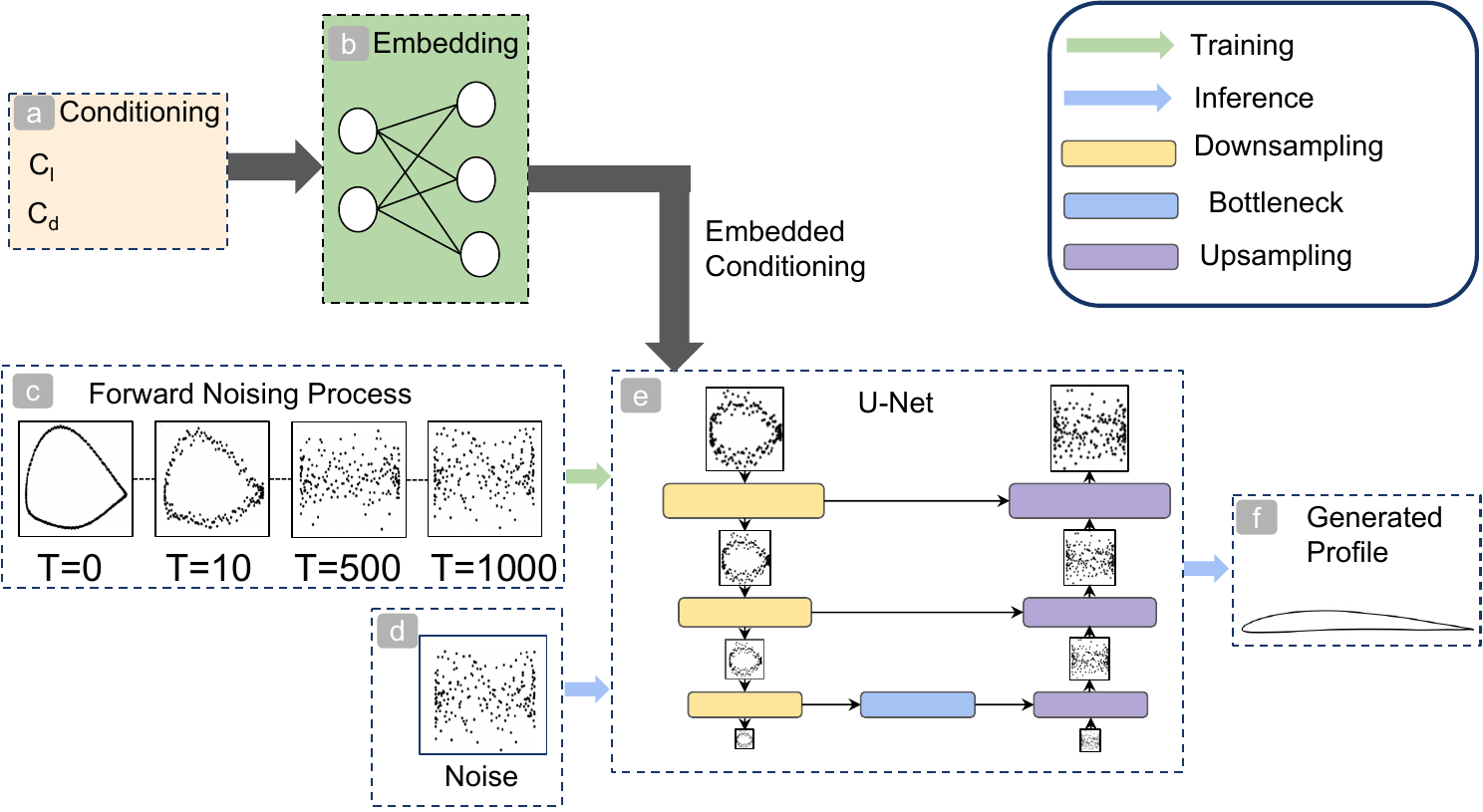}
    \caption{Overview of the Airfoil Diffusion Model:
    \newline 
    a) Airfoil conditioning information such as the coefficient of lift (\(C_l\)), coefficient of drag (\(C_d\)), maximum thickness, and maximum camber is passed through b) conditional embedding multi-layer perceptron (MLP) layers. The embedded conditioning information is passed to e) the denoising U-Net. During training, samples are progressively distorted to approximate Gaussian noise, as shown in c). In the inference phase, a noise vector d) along conditioning value embedding are input to the U-Net e), which outputs the conditionally generated airfoil profile in f). 
    }
    \label{model_overview}
\end{figure}

\section{Related Work}
We propose a novel approach to airfoil design using conditional diffusion models that build upon existing methodologies in geometric parameterization and deep learning.

\subsection{Geometry Parameterization}

Traditional airfoil design has predominantly relied on computational simulations and fixed design parameters, which can restrict the exploration of innovative aerodynamic shapes \cite{Barrett2006Airfoil}. These conventional methods, though effective in specific scenarios, often limit the ability to fully explore the expansive design space of aerodynamic structures \cite{Morris2008Wing}. This limitation arises primarily from the need for predefining the design space using specific shape parameters, such as camber and thickness, which can constrain the range of possible geometries that can be considered during the airfoil design process.

To address these limitations, there has been a shift towards advanced machine learning techniques that allow for more flexible and efficient exploration of the design space. These approaches, including techniques like Bézier and B-spline parameterizations, offer greater control over airfoil geometry by representing shapes with a set of control points that can be adjusted to generate airfoil coordinate and continuous surfaces. Such methods have been integrated into optimization frameworks to enhance the precision of geometric design while maintaining smoothness and manufacturability \cite{Zhang2020BezierOptimization, Mukesh2021BSpline}. However, while these traditional geometric parameterizations provide a level of flexibility, they still require significant manual intervention and are often limited in handling highly complex shapes, especially in high-dimensional design spaces.

The introduction of data-driven methods, particularly deep learning models, has further advanced the field of geometry parameterization. By learning from large datasets of existing airfoil shapes, these models can generate new designs that meet specific aerodynamic performance criteria without being constrained by traditional shape parameters. Variational Autoencoders (VAEs) and Generative Adversarial Networks (GANs) have been particularly impactful in this domain. For example, the work by Li et al. \cite{Li2020GeometricGAN} demonstrated the application of GANs to generate a diverse set of airfoil shapes by encoding them into a latent space, thereby enabling a more extensive exploration of the design space compared to traditional methods.

Moreover, Diffusion Models, as explored in recent studies, represent a significant advancement in airfoil geometry parameterization. These models operate within a learned latent space, allowing for the generation of airfoil shapes with fewer training samples and maintaining high-quality outputs \cite{Wei2023Diffusion}. The ability of Diffusion Models to integrate specific design constraints directly into the generative process provides a powerful tool for generating aerodynamic shapes in a data-efficient manner.

As the field continues to evolve, hybrid approaches that combine traditional geometric parameterization with advanced machine learning techniques are emerging. These methods aim to leverage the strengths of both worlds—preserving the airfoil coordinateness and manufacturability of traditional parameterizations while exploiting the flexibility and data-driven insights of deep learning models. The ongoing research in this area focuses on refining these techniques to accommodate increasingly complex aerodynamic and geometric constraints, ultimately pushing the boundaries of what is possible in airfoil design.

\subsection{Deep Learning Approaches}

Generative models, including Variational Autoencoders (VAEs) and Generative Adversarial Networks (GANs), have shown considerable promise in advancing airfoil design by overcoming the limitations of traditional design methods. For instance, the Airfoil GAN model \cite{Wang2021Airfoil} encodes existing airfoil shapes into latent vectors, enabling the generation of novel airfoils with specific aerodynamic properties, without relying on predefined design parameters. This approach not only facilitates the exploration of high-dimensional design spaces but also integrates a genetic algorithm for optimizing aerodynamic performance, allowing the synthesized airfoils to evolve towards desired properties. Moreover, the model's ability to generate airfoil coordinate and realistic airfoils directly from the latent space, coupled with its interpretability through feature clustering, sets it apart from traditional techniques. In comparison, the use of free-form deformation generalized adversarial networks (FFD-GAN) \cite{Chen2021Deep} has advanced the parameterization of three-dimensional aerodynamic shapes, yet it often requires predefined design formulations, limiting its flexibility in exploring novel geometries. The Airfoil GAN's potential for application to other aerodynamic and engineering domains further underscores its significance in the field of design optimization.

Despite their potential, GAN-based methods often face challenges related to training stability and the need for large datasets. Addressing these issues, DiffAirfoil \cite{DiffAirfoil} introduces a novel approach that leverages a diffusion model within a learned latent space, specifically tailored for aerodynamic shape optimization. This model significantly enhances data efficiency, requiring substantially fewer training geometries compared to GANs, while still generating high-quality and diverse airfoil samples. The diffusion process not only ensures stable training and robust sampling, even under data-scarce conditions, but also mitigates the common pitfalls associated with adversarial training, such as mode collapse \cite{mangalam2021overcoming, bayat2023a}.

Moreover, DiffAirfoil's conditional variant offers a distinct advantage in generating airfoils that meet specific geometric constraints without the need for retraining or model adjustments, thereby providing greater flexibility and efficiency in the design process. This capability is particularly valuable in scenarios where rapid design iterations are necessary, as it allows for the direct incorporation of user-defined constraints, such as maximum thickness or area, into the generated airfoil profiles.

However, DiffAirfoil’s approach is inherently tied to the use of a latent space for diffusion, where the model operates on latent vectors derived from an autodecoder. This methodology necessitates a template airfoil profile as a starting point, with the generated airfoils being deformed versions of this template, shaped according to the sampled latent vector. Although this approach is effective for optimizing and blending existing designs, it is fundamentally constrained by the reliance on predefined templates, which may limit the model’s ability to fully explore the design space and discover novel and unconventional airfoil geometries.

In contrast, our work advances deep learning methodologies by proposing a conditional diffusion model that operates directly in the airfoil geometric space, without the need for latent space transformations or template geometries. By avoiding the constraints of template-based generation, our model focuses on the exploration of the design space rather than the optimization of existing designs. This approach enables our model to learn a comprehensive representation of the entire training dataset and generates airfoils that are not bound by the limitations of pre-existing shapes. Consequently, our model is more generalizable and capable of exploring a broader design space, offering greater flexibility in generating innovative airfoil geometries that might not be achievable through latent space diffusion models like DiffAirfoil.

\section{Methodology}
This section outlines the method for developing the airfoil diffusion model. The first subsection details the airfoil training dataset, including data processing for model training. The second subsection explores the method for obtaining airfoil performance measurements. The third subsection presents the background theory of diffusion models and the formulation of the airfoil diffusion model.

\begin{figure}[h]
    \centering
    \includegraphics[width=\linewidth]{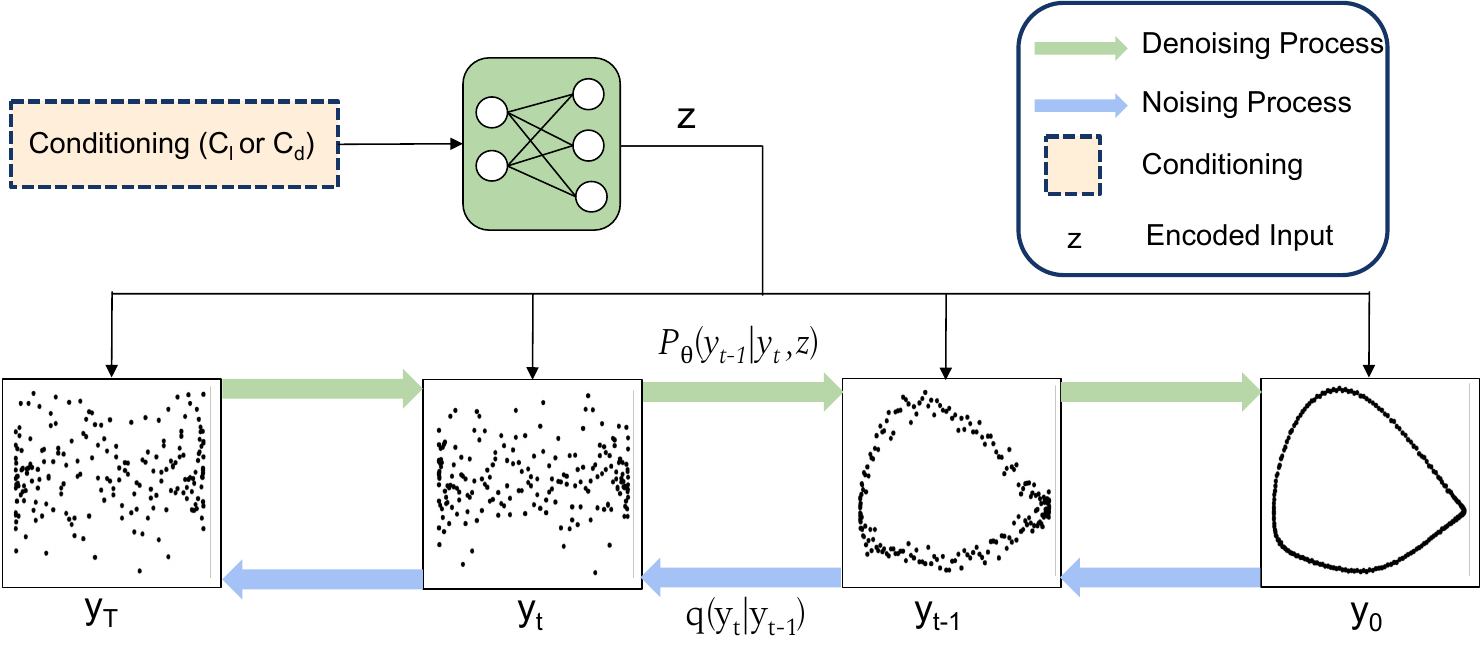}
    \caption{The conditioning input, such as \(C_d\) or \(C_l\) is fed through several multilayer perceptron layers to generate a context vector \(z\) for conditioning the diffusion model. In the forward process, the sample y coordinates are subjected to noise for \(T\) timesteps, gradually approaching gaussian noise. During the backward process, the model learns to estimate the added noise at each timestep \(t\) given conditioning \(z\) and \(y_t\).}
    \label{diffusion_process}
\end{figure}

\subsection{Dataset}
The UIUC Airfoil Database \cite{Selig1996}, hosted by the University of Illinois Urbana-Champaign, contains approximately 1600 airfoils, each represented by detailed coordinates.

Airfoil profiles in the UIUC database consist of varying numbers of (x, y) coordinate pairs. However, neural networks require uniform input dimensions, necessitating a standardized number of data points across all samples. To address this, we repanelized the airfoil profiles using Aerosandbox \cite{aerosandbox}, interpolating each sample to a consistent set of 200 points via cubic splines. This approach preserves the airfoil's shape and characteristics while ensuring uniform data representation.

The airfoils are normalized such that x-coordinates range from 0 to 1 and y-coordinates from -1 to 1. After repanelization, all samples share identical x-coordinates, effectively reducing the problem dimensionality from two to one dimension, which simplifies both training and generation processes.

The y-coordinate values are structured as follows: The first 100 points, corresponding to x-values decreasing from 1 to 0, represent the upper surface of the airfoil, while the last 100 points, corresponding to x-values increasing from 0 to 1, represent the lower surface. During training, we organize these y-values into two channels to help the model differentiate between the upper and lower surfaces of the airfoils.

\subsection{Obtaining Aerodynamic Coefficients}

The performance of the airfoil is evaluated based on the lift and drag produced by the airfoil \cite{koning2023improved}. Lift is produced by the change in air velocity as it passes over the airfoil, while drag is created by skin friction and pressure changes due to flow separation and shocks \cite{raymer2006aircraft}. 

\subsubsection{Airfoil Geometry}

The lift and drag produced by an airfoil are directly related to the shape of its profile \cite{raymer2006aircraft}. The chord length is the straight line distance from the leading edge to the trailing edge of the airfoil. The camber represents the curvature of the profile and is defined as the line passing through the center of the airfoil, equidistant from the upper and lower surfaces. The airfoil thickness ratio is an additional feature of airfoils and is derived from the chord length \(c\) and maximum thickness \(t\) of the airfoil \cite{raymer2006aircraft}:
\begin{align*}
    \text{Airfoil Thickness Ratio} &= \frac{t}{c}
 \end{align*}

The angle of attack \(\alpha\) is the angle between the chord line of the airfoil and the vector representing the relative motion of the airfoil through the air. This angle influences the lift and drag experienced by the airfoil by affecting the airflow over its surface \cite{gracey1958summary}.

To evaluate the aerodynamic performance of both the training and the synthesized airfoil profiles, we utilized NeuralFoil \cite{neuralfoil}, a neural network-based tool designed to predict aerodynamic coefficients. NeuralFoil provides an efficient and accurate means of obtaining the coefficient of lift (\(C_l\)) and the coefficient of drag (\(C_d\)) for a given airfoil geometry.

\subsubsection{NeuralFoil Overview}

NeuralFoil leverages a deep learning model trained on a comprehensive dataset of airfoil shapes and their corresponding aerodynamic coefficients. This model is able to predict \(C_l\) and \(C_d\) based on the input airfoil coordinates. While some skepticism may arise regarding whether a neural network based model can adhere to physical principles, NeuralFoil addresses this concern by using a physics-informed neural network (PINN) architecture trained on tens of millions of XFOIL simulations \cite{neuralfoil}. This ensures that the model respects aerodynamic laws while providing rapid predictions. Furthermore, NeuralFoil has been benchmarked against XFOIL, achieving accuracy within a few percent of XFOIL's predictions while offering significant computational speedups \cite{neuralfoil}. Additionally, NeuralFoil provides an ``analysis confidence" output for each prediction, quantifying uncertainty and ensuring robust design optimization \cite{neuralfoil}.

\subsubsection{Process for Obtaining Coefficients}

For each airfoil profile in our dataset, we followed these steps to obtain the aerodynamic coefficients.

\begin{enumerate}
    \item Preparation of Input Data: Each airfoil profile, represented by 200 points along the \(x\) and \(y\) coordinates, was formatted as required by NeuralFoil.
    \item Prediction Using NeuralFoil: The prepared airfoil data was fed into NeuralFoil, which predicted the corresponding \(C_l\) and \(C_d\) values.
    \item Post-Processing: The predicted \(C_l\) and \(C_d\) values were recorded and used for further analysis. These coefficients were essential for filtering the synthesized airfoils and evaluating their performance characteristics.
\end{enumerate}

\subsection{Denoising Diffusion Probabilistic Models (DDPMs)}

DDPMs \cite{Croitoru2022Diffusion, Ho2021Cascaded, ho2020denoising, ogoke2023inexpensive, Yang2022Diffusion, jadhav2024generative} are a class of probabilistic models inspired by nonequilibrium thermodynamics \cite{sohldickstein2015deep}. These models learn to reverse the iterative corruption process applied to the training data. The training procedure consists of two main phases:

\begin{enumerate}
    \item The forward diffusion process.
    \item The reverse diffusion process.
\end{enumerate}

\subsubsection{Forward Process}
During the forward diffusion process, the training data are gradually corrupted until they approximate Gaussian noise. This corruption is achieved using a Markovian process \cite{ho2020denoising, Ho2021Cascaded, jadhav2023stressd, jadhav2024generative, ogoke2023inexpensive, Croitoru2022Diffusion}, where the probability of each event depends only on the state of the previous event. In other words, the next future state depends only on the current state \cite{ho2020denoising}. In our Airfoil diffusion model, the data distribution at step $t$ is given by:

\begin{align}
    q(y_t|y_{t-1}) &= \mathcal{N}\left(y_t;\sqrt{1-\beta_t}\cdot y_{t-1}, \beta_t\cdot\textbf{I}\right)
\end{align}
Where:
\begin{align*}
    t &\in \left(0, 1, \dots, T\right)
    \\
    \beta_1, \beta_2, \dots, \beta_T &\in \left[0, 1\right)
\end{align*}

Here, $q(y_0)$ represents the original data distribution. As $t \rightarrow T$, $q(y_t)$ approaches a Gaussian distribution. The variance schedule $\beta_t$ is a hyperparameter. $\mathcal{N}(y; \mu, \sigma)$ represents the normal distribution with mean $\mu$ and covariance $\sigma$ that produces $y$ \cite{Croitoru2022Diffusion}.

While the Markovian process is defined step-by-step, we improve computational efficiency by leveraging the following properties:
\begin{enumerate}
    \item Each step in the process adds Gaussian noise.
    \item The sum of Gaussian distributions is also Gaussian \cite{degroot2014probability}.
\end{enumerate}
Substituting:
\begin{align}
    \alpha_t &= 1-\beta_t \\
    \Bar{\alpha_t} &= \prod_{s=1}^t\alpha_s \\
    \epsilon &\sim \mathcal{N}(0,1)
\end{align}
We define our sample and data distribution as follows:
\begin{align}
    y_t &= \sqrt{\Bar{\alpha}_t}y_0 + \sqrt{1-\Bar{\alpha}_t}\epsilon \\
    q(y_t|y_0) &= \mathcal{N}\left(y_t;\sqrt{\Bar{\alpha}_t}y_0, (1-\Bar{\alpha}_t)\textbf{I}\right)
\end{align}

\subsubsection{Reverse Process}
Starting from a random sample from the distribution defined in the forward process, we generate new airfoil samples within the original training data distribution \cite{ho2020denoising, Ho2021Cascaded, Croitoru2022Diffusion}. This stepwise process is represented by the following distribution:

\begin{align}
    p_\theta\left(y_{t-1}|y_t\right) &= \mathcal{N}\left(y_{t-1};\mu_\theta(y_t,t), \Sigma_\theta(y_t,t)\right)
\end{align}

By substituting equations (2-4), we obtain our reconstructed data distribution and individual sample:
\begin{align}
    p_\theta(y_{t-1}|y_t) &= \mathcal{N}\left(y_{t-1};\frac{1}{\sqrt{\alpha_t}}\left(y_t-\frac{\beta_t}{\sqrt{1-\bar{\alpha}_t}}\epsilon_\theta(y_t,t)\right), \beta_t\textbf{I}\right) \\
    y_{t-1} &= \frac{1}{\sqrt{\alpha_t}}\left(y_t - \frac{\beta_t}{\sqrt{1-\bar{\alpha}_t}}\epsilon_\theta(y_t,t)\right)+ \sqrt{\beta_t}\epsilon
\end{align}

\subsubsection{Airfoil Diffusion}
The airfoil diffusion model generates novel airfoils by transforming random one-dimensional vectors through a trained neural network. This model allows for the generation of an arbitrary number of novel airfoils, limited only by the available computational power.

The forward diffusion process gradually adds Gaussian noise to an airfoil sample from the dataset, based on a predefined noise schedule $q(y_t|y_{t-1})$, until the sample approximates Gaussian noise. In contrast, the reverse process involves gradually removing the noise from a noised sample to reconstruct an approximation of the original airfoil. 

This denoising process is performed by a neural network trained to predict the noise added at each time step. By learning to predict the added noise, the network effectively maps random vectors of length 200 to various airfoil geometries.

To predict the noise added to the samples, we used the U-net architecture \cite{ho2020denoising, ronneberger2015unetconvolutionalnetworksbiomedical, ogoke2023inexpensive}. The U-net functions similarly to an autoencoder, where complex data is encoded into a latent representation and then decoded back. In our application, a noised sample is passed to the U-net along with timestep information. The U-net encodes the sample into a latent space and, using skip connections \cite{ronneberger2015unetconvolutionalnetworksbiomedical}, retains information to aid in the decoding process. The output of the U-net is the predicted noise vector that was added to the input sample.

\subsection{Conditional Diffusion}
Standard diffusion models are effective at generating novel airfoil samples but often lack mechanisms to ensure that these samples meet specific geometric or performance-based criteria. To address this, we incorporate a conditional diffusion model that steers the generation process toward desired outcomes based on conditioning parameters, such as geometric features or aerodynamic performance metrics \cite{ho2022classifierfreediffusionguidance, hu2024drugdiscoverysmilestopharmacokineticsdiffusion, jadhav2023stressd,jadhav2024generative}.

\subsubsection{Conditioning Mechanism}
The conditional diffusion model integrates additional information, such as geometric constraints or aerodynamic coefficients, directly into the generative process. The conditioning data (e.g., lift coefficient \(C_l\), maximum thickness, or maximum camber) is processed through linear transformations to map it into a high-dimensional embedding space. This is followed by a ReLU activation function.

The resulting conditioning embedding is fused with timestep information using a bilinear layer, combining both into a single representation that captures the progression of the diffusion process and target constraints.

\subsubsection{Conditional U-Net Architecture for 1D Data}

We employ a Conditional U-Net architecture to incorporate conditioning information during airfoil generation. The U-Net’s encoder-decoder structure, enhanced with skip connections, is well-suited for retaining spatial information at various scales. In our case, the U-Net is adapted for 1D, two-channel input data (e.g., airfoil coordinates), replacing the traditional 2D convolutions with 1D convolutions for downsampling and upsampling operations.

The input data is first processed by an initial 1D convolutional layer to match the network's dimensionality requirements. The downsampling path consists of multiple stages, each containing ResNet blocks, linear attention layers, and downsampling operations (either strided or learned convolutions). These layers progressively reduce the spatial dimensions while increasing feature channels. The upsampling path mirrors this structure but progressively increases spatial dimensions while reducing feature channels. The final output is produced by residual blocks and a 1D convolutional layer.

To effectively integrate conditioning information (such as desired aerodynamic coefficients \(C_l\) and \(C_d\)), we fuse the conditioning data with timestep embeddings at each stage of the U-Net. Time step information is encoded via sinusoidal embeddings processed through a multi-layer perceptron (MLP), while conditional inputs are processed through linear layers and bilinear layers to generate conditional embeddings. These embeddings are combined with time embeddings and injected into ResNet blocks throughout both downsampling and upsampling paths.

Mathematically, input data is encoded into a latent representation via the U-Net’s downsampling path. The combined time and conditional embeddings are used to condition ResNet blocks across the network, ensuring that generated outputs satisfy specified conditions. The latent representation is then decoded back into structured data through upsampling layers, guiding the model toward airfoil designs that meet target specifications.

%
%
%

\section{Results and Discussion}

In this section, we explore the performance and capabilities of the Airfoil Diffusion model, trained on the UIUC airfoil database, through both unconditional and conditional generation. The results are analyzed based on geometric properties, aerodynamic performance, and the novelty of the generated airfoil profiles.

\subsection{Generated Profile Smoothing}

To address the lack of guaranteed smoothness in airfoils generated by the diffusion model, we implement a smoothing framework based on B-spline interpolation, which offers a smooth approximation of the airfoil shape \cite{bsplines, DIERCKX1975165}. This is achieved using the \texttt{BSpline} and \texttt{splrep} functions provided by the SciPy Python library \cite{2020SciPy-NMeth}.

To ensure compatibility with this method, we preprocess the airfoil coordinates by separating the upper and lower surfaces and reordering the points to have increasing $x$ values. For the upper surface, where $x$ values are typically decreasing, we reorder the points as:
\begin{align*}
\{(x_i, y_i)\}_{i=1}^{n} \rightarrow \{(x_{\sigma(i)}, y_{\sigma(i)})\}_{i=1}^{n},
\end{align*}
where $\sigma$ is a permutation such that $x_{\sigma(1)} < x_{\sigma(2)} < \dots < x_{\sigma(n)}$.

After smoothing, the B-spline representation is restored to the original orientation to preserve the original data format while achieving a smoother representation.

\subsection{Shape Similarity Measurement}

To evaluate the similarity between the generated airfoils and the training dataset, we utilized Principal Component Analysis (PCA), a dimensionality reduction technique that transforms high-dimensional data into a lower-dimensional space while preserving the most significant modes of variation \cite{yonekuraPCA, mackiewicz1993principal}. PCA computes linear combinations of the original variables, called principal components, which are uncorrelated and ranked by the variance they explain in the data.

The PCA process begins by centering the observation matrix $\mathbf{X}$, where each row represents an airfoil, by subtracting the mean of each feature. The variance-covariance matrix $\mathbf{S}$ is then constructed, and the principal components $\mathbf{V}$ are calculated as:

\[
\mathbf{V} = \mathbf{A}^\top \mathbf{X},
\]

where $\mathbf{A}$ is a matrix of eigenvectors of $\mathbf{S}$, and the eigenvalues $\lambda_i$ represent the variance captured by the corresponding principal components. For this study, PCA was applied to the training dataset to project airfoil shapes into a reduced-dimensional space. The same transformation was then applied to the generated airfoils, enabling direct comparisons in this space.

\subsection{Unconditional Generation}

\begin{figure}[H]
\centering
\includegraphics[width=\linewidth]{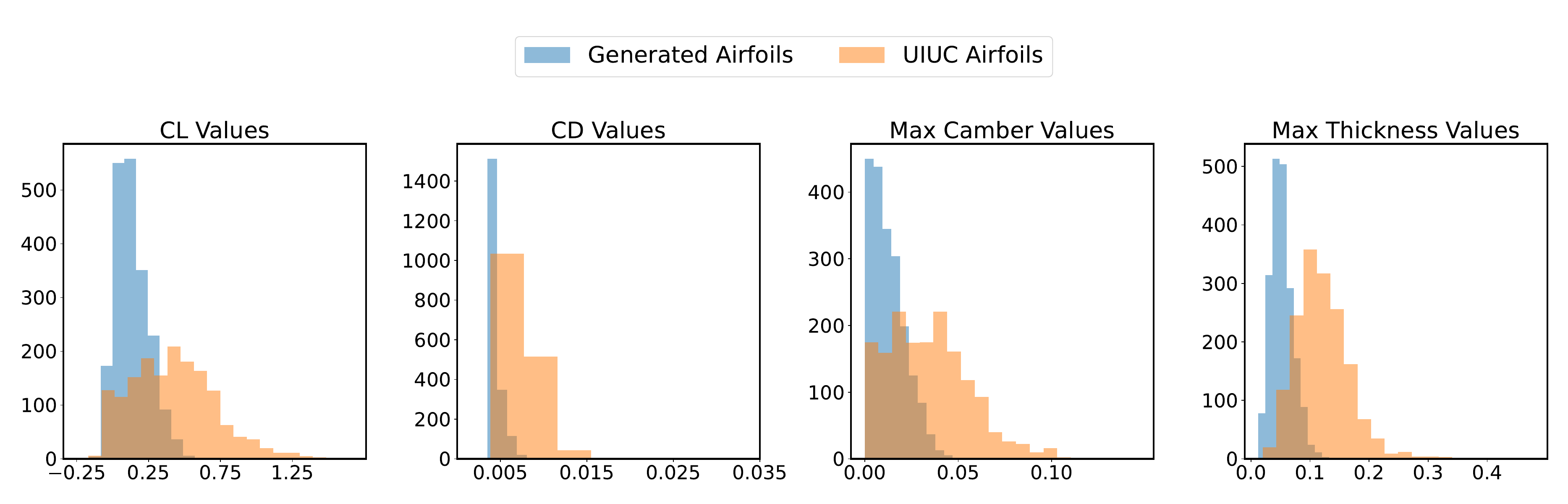}
\caption{Histograms of \(C_l\), \(C_d\), maximum camber, and maximum thickness for the UIUC (orange) and generated (blue) samples. The generated samples exhibit narrower distributions compared to the UIUC data, with their means slightly shifted to the left of the UIUC distributions.}
\label{gen_uiuc_hist}
\end{figure}

To assess the capabilities of the unconditional Airfoil Diffusion model, 2000 random samples were generated by feeding random vectors into the model. The output airfoil profiles were evaluated based on their geometric and aerodynamic properties, and the resulting distributions of lift coefficient (\(C_l\)), drag coefficient (\(C_d\)), maximum camber, and maximum thickness were compared with those from the UIUC dataset. Figure \ref{gen_uiuc_hist} presents histograms of these distributions.

The generated airfoils exhibit a narrower range of \(C_l\) values, ranging from -0.1 to 0.5, compared to the broader range of -0.2 to 1.25 observed in the UIUC dataset. Similarly, the \(C_d\) values for the generated airfoils span from 0.001 to 0.009, whereas the UIUC dataset shows a range from 0.001 to 0.015. In terms of geometric properties, the generated airfoils have maximum camber values between 0 and 0.05 and maximum thickness between 0.01 and 0.13, which are more constrained compared to the UIUC dataset (0 to 0.1 for camber and 0.01 to 0.35 for thickness).

The narrower distributions and the leftward shift in means for all properties suggest that the model tends to generate airfoils that are more conservative and less extreme than those in the UIUC dataset. Specifically, the underrepresentation of high \(C_l\) and \(C_d\) values, as well as extreme thickness and camber, indicates that the model may favor the generation of more typical airfoil shapes rather than exploring the full range of geometries present in the training set. This behavior could be attributed to the inherent biases in the training data distribution or the diffusion process, which might favor the generation of profiles closer to the dataset's mean.

\subsection{Conditional Generation}
\begin{figure}[H]
    \centering
    \includegraphics[width=\linewidth]{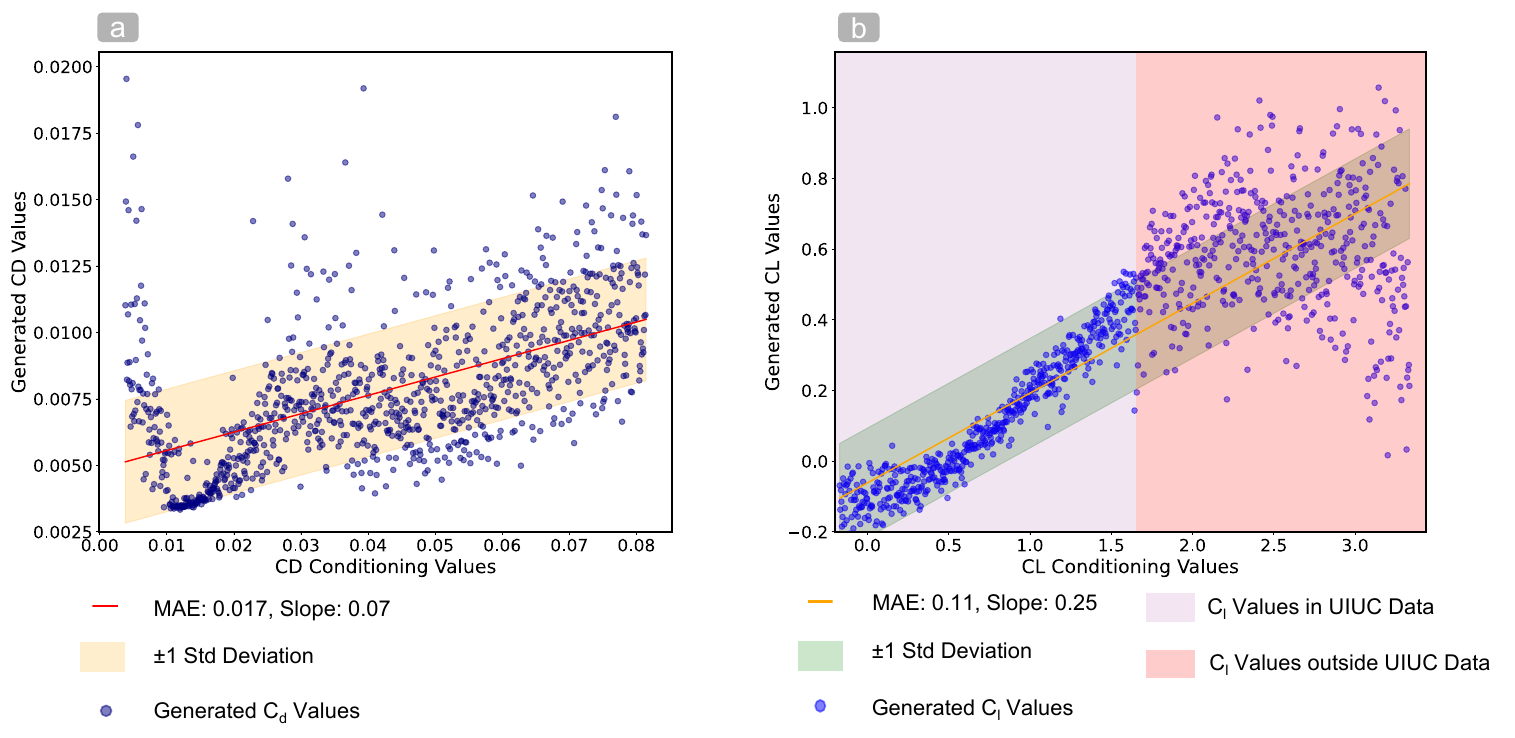}
    \caption{ Scatter plots of the relationship between input conditioning values for \(C_d\)  and \(C_l\) in (a) and (b), respectively. Also shown are the lines of best fit, along with the visual representation of the standard deviation from the line of best fit. In (b), the purple region depicts conditioning on values of \(C_l\) seen in the training dataset, and the red region shows conditioning on values not seen in the UIUC dataset.}
    \label{conditioning}
\end{figure}

The conditional diffusion model was evaluated by generating 1000 samples conditioned on specific values of \(C_l\) and \(C_d\). Two separate models were trained: one conditioned on \(C_d\) and the other on \(C_l\). The input conditioning values were linearly spaced between the minimum and maximum values of \(C_d\) and the minimum and twice the maximum of \(C_l\) in the UIUC dataset to test the model's potential to generate samples with greater \(C_l\) values.

Analysis of the relationship between the conditioned and generated values was conducted by plotting scatter plots of the input conditioning values against the NeuralFoil-evaluated results (Fig. \ref{conditioning} (b), (a)). The mean absolute error (MAE) was calculated to quantify the accuracy of the conditioning. The \(C_d\)-conditioned model achieved an MAE of 0.0018, whereas the \(C_l\)-conditioned model had a higher MAE of 0.04. The slope of the best-fit line for the \(C_d\)-conditioned model was 0.07, compared to 0.25 for the \(C_l\)-conditioned model, indicating that the \(C_d\)-conditioned model's outputs were less sensitive to the input conditioning values. As the \(C_l\) conditioning values exceeded the values seen in the UIUC samples, the \(C_l\) conditioned model generations exhibited a greater standard deviation, however the model still tended to produce airfoils with greater \(C_l\) values corresponding to increasing conditioning values. This effect is likely due to passing conditioning values to the model that were not seen in training samples.

The range of output \(C_d\) values was 0.0033 to 0.0226, and the range of output \(C_l\) values was -0.3319 to 1.09. These outputs were narrower compared to the input conditioning ranges of 0.038 to 0.0814 for \(C_d\) and -0.1701 to 3.21 for \(C_l\). The maximal \(C_l\) value observed in the UIUC dataset is approximately 1.25; however, this is an outlier. The majority of airfoils in the dataset exhibit \(C_l\) values between 0.0 and 0.75, suggesting that the model is sufficiently capable of representing most samples in the dataset, excluding the most extreme cases.

To approximate a one-to-one correspondence between input conditioning values and output airfoil performance coefficients, the input values can be scaled by a constant factor equal to the inverse slope of the line of best fit. This adjustment ensures alignment, provided the desired conditioning values remain within the bounds supported by the model.

\subsubsection{Aerodynamic Performance}

Evaluating the aerodynamic performance of the generated airfoils is crucial, extending beyond merely targeting specific \(C_d\) and \(C_l\) values. The lift-to-drag ratio (\(C_l/C_d\)) is a key metric that provides a comprehensive assessment of an airfoil's efficiency \cite{Wang2021Airfoil, akram2021aerodynamic}.

\begin{figure}[H]
    \centering
    \includegraphics[width=\linewidth]{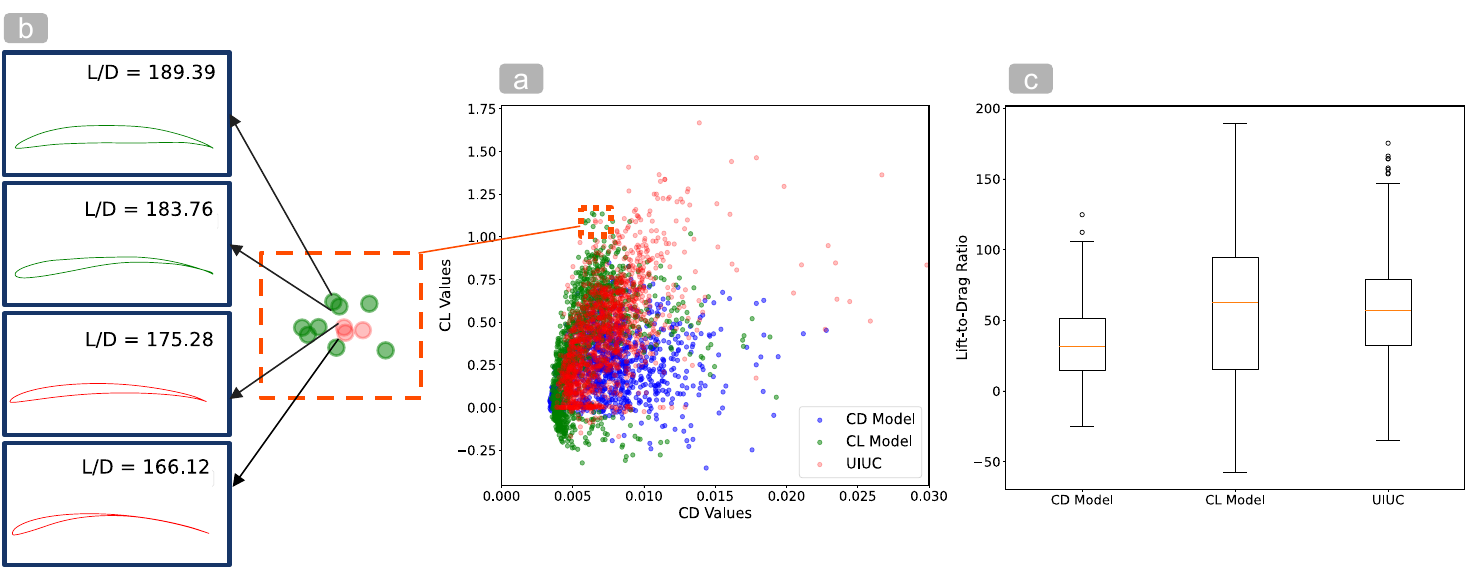}
    \caption{a) Scatter plot of the \(C_d\) (x-axis) and \(C_l\) (y-axis) values from NeuralFoil evaluations for both the \(C_d\)-conditioned (blue points) and \(C_l\)-conditioned (green points) models. The \(C_d\)-conditioned model spans a broader range of \(C_d\) values, while the \(C_l\)-conditioned model spans a broader range of \(C_l\) values, with a bias towards lower \(C_d\) values. b) Airfoils with the greatest lift-to-drag ratios from the \(C_l\)-conditioned model (green) and UIUC dataset (red). The top UIUC airfoils have elongated shapes with minimal thickness and varying degrees of camber, while the \(C_l\)-conditioned model generates airfoils that similarly exhibit reduced thickness and differing camber profiles. The \(C_l\)-conditioned airfoils generally achieve higher lift-to-drag ratios with slightly more thickness and less camber compared to the UIUC samples. c) Box plots of the lift-to-drag ratio for the \(C_d\)- and \(C_l\)-conditioned models and the UIUC dataset. The \(C_l\)-conditioned model exhibits a wider range of lift-to-drag ratios and produces samples with greater lift-to-drag ratios than those seen in the UIUC dataset.}
    \label{cl_vs_cd}
\end{figure}

We evaluated the performance of generated airfoils using the NeuralFoil library \cite{neuralfoil}, with angle of attack set to 0 degrees, Reynolds number of $1\times 10^6$ and mach number 0.0. Figure \ref{cl_vs_cd}(a) presents a scatter plot of the evaluated \(C_d\) and \(C_l\) values for both the \(C_d\)-conditioned and \(C_l\)-conditioned models. The \(C_d\)-conditioned model exhibits a broader range of \(C_d\) values, demonstrating its ability to generate airfoils with diverse drag characteristics. However, it produces a narrower range of \(C_l\) values, indicating more constrained lift performance. Conversely, the \(C_l\)-conditioned model generates a broader spectrum of \(C_l\) values, showcasing its capacity to explore varied lift characteristics, while producing airfoils with a narrower range of \(C_d\) values, favoring lower drag. This bias towards lower drag in the \(C_l\)-conditioned model is advantageous for maximizing the lift-to-drag ratio, making it particularly effective in producing efficient airfoil designs.

Figure \ref{cl_vs_cd}(b) presents the airfoils with the greatest lift-to-drag ratios from the \(C_l\)-conditioned model (green) and UIUC dataset (red). The top two UIUC airfoils have lift-to-drag ratios of 175.28 and 166.12. Both profiles are characterized by elongated shapes with minimal thickness. The profile with a lift-to-drag ratio of 166.12 exhibits significant camber and an extremely thin profile starting from approximately one-third of the chord length to the trailing edge. The profile with a lift-to-drag ratio of 175.28 is slightly thicker, with less camber, tapering sharply in thickness towards the trailing edge. Similarly, the top two airfoils from the \(C_l\)-conditioned model exhibit relatively little thickness. The profile with a lift-to-drag ratio of 183.76 is slightly thicker than the UIUC sample with a ratio of 166.12, with extremely thin thickness starting at approximately the last third of the chord length, and slightly less camber. The profile with a lift-to-drag ratio of 189.39 is distinct, with the greatest thickness among the samples and less overall camber. The lower half of this profile has a relatively straight section for the middle two-thirds of the chord, with slight camber at the leading and trailing edges. The upper half has moderate camber with a slight increase at the leading and trailing edges.

Figure \ref{cl_vs_cd}(c) compares the lift-to-drag ratios for the \(C_d\)- and \(C_l\)-conditioned models against the UIUC dataset through box plots. The \(C_d\)-conditioned model demonstrates the narrowest range of lift-to-drag ratios, indicating a more consistent yet potentially less exploratory design space. In contrast, the \(C_l\)-conditioned model shows a wider range of lift-to-drag ratios, with some generated airfoils exceeding the maximum ratios observed in the UIUC dataset. This result highlights the \(C_l\)-conditioned model's ability to generate airfoils with superior aerodynamic performance, surpassing the capabilities of existing designs within the training data. These outcomes underscore the model's potential in advancing airfoil design by pushing aerodynamic efficiency beyond conventional limits.

From the lift-to-drag ratio comparison between the conditioned models and the UIUC dataset, it is evident that the \(C_l\)-conditioned model is capable of generating airfoils with performance characteristics that surpass those found in the training samples. This finding highlights the applicability of conditional airfoil diffusion models in real-world aerodynamic design, offering the potential to enhance the efficiency of future airfoil designs.

\subsubsection{Design Space Analysis}

\begin{figure}[H]
    \centering
    \includegraphics[width=\linewidth]{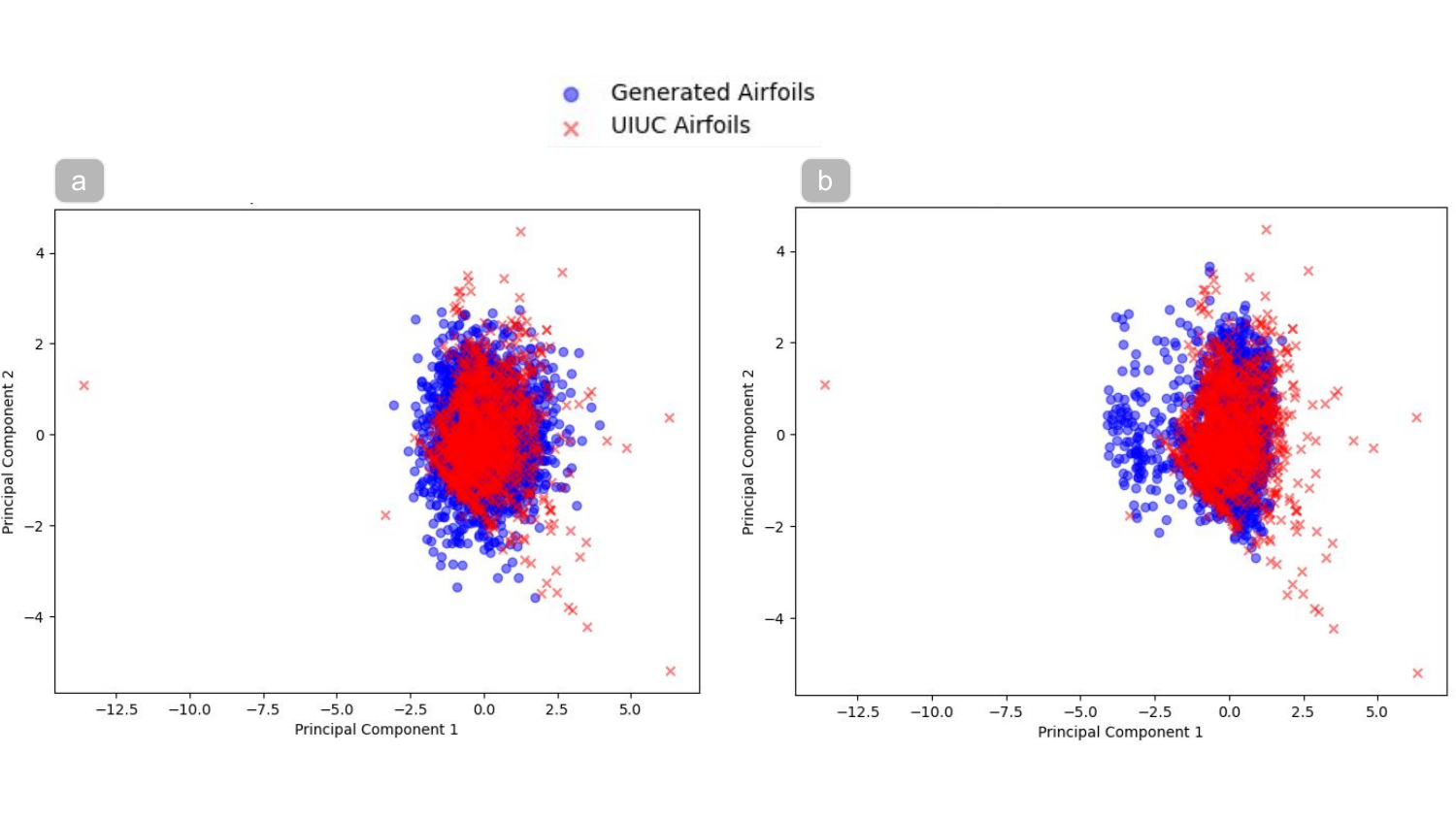}
    \caption{PCA projections of the UIUC dataset samples compared with the principal components of the \(C_l\)-conditioned model (Figure a) and the \(C_d\)-conditioned model (Figure b). The principal components of both models align closely with the most frequently occurring projections in the UIUC dataset. The \(C_l\)-conditioned model projections encompass the UIUC projections uniformly, while the \(C_d\)-conditioned model projections also encompass the dataset but are noticeably shifted to the left relative to the center of the UIUC distribution.
}
    \label{PCA}
\end{figure}

To assess the overlap and exploration of the design space, we applied PCA to both the training dataset and the generated airfoils conditioned on \(C_l\) and \(C_d\). Figure~\ref{PCA} shows the PCA projections of the UIUC dataset alongside the generated samples.

From Figure~\ref{PCA}, the \(C_l\)-conditioned model (Figure~\ref{PCA}(a)) is centered around the most frequently occurring projections of the UIUC dataset, with a slight expansion in the space covered. In contrast, the \(C_d\)-conditioned model (Figure~\ref{PCA}(b)) demonstrates a shift to the left, indicating an exploration of different characteristics of the design space compared to the \(C_l\)-conditioned model.

This result suggests that the airfoil diffusion model adheres to the design space defined by the training dataset, ensuring physically feasible airfoil designs. Simultaneously, the model exhibits the capability to generate novel geometries within this realistic design space, offering potential for discovering new airfoil profiles with favorable performance characteristics. This feature underscores the utility of the diffusion model in assisting airfoil design exploration and innovation.

\section{Conclusion}

In this paper, we presented a novel diffusion-based framework for airfoil design, leveraging the UIUC airfoil database to train a model capable of generating new designs within the training sample design space. By conditioning the generation process on specific aerodynamic coefficients such as lift (\(C_l\)) and drag (\(C_d\)), our model demonstrated the ability to produce airfoils with desirable aerodynamic and geometric criteria not seen in training samples.

Our experiments revealed that the diffusion model is effective not only in generating airfoils that align well with the overall distribution of the training dataset but also in producing designs that push the boundaries of conventional airfoil geometries. The unconditional generation results showed a tendency towards conservative designs, while the conditional generation results highlighted the model's flexibility in achieving specific performance targets, with the \(C_l\)-conditioned model excelling in generating airfoils with superior lift-to-drag ratios.

One of the key strengths of our approach lies in its capacity to produce novel airfoil geometries, as evidenced by the PCA reduction analysis. The generated airfoils, while grounded in the characteristics of the training data, exhibit moderate novelty, indicating the model's potential for discovering innovative aerodynamic shapes. Importantly, the generated designs remain within a realistic and aerodynamically viable space, ensuring their practical applicability.

In summary, the diffusion model-based framework we introduced represents a powerful tool for exploring the design space and discovering new, high-performance airfoil shapes. Future work could explore extending this methodology to three-dimensional aerodynamic shapes, integrating multiple conditioning parameters within a single model, and applying this generative approach to other domains where innovative design solutions are critical.

\section*{Declarations}

\begin{itemize}
\item Funding:
No funding was received to assist with the preparation of this manuscript.
\item Competing interests:
The authors did not receive support from any organization for the submitted work.
\item Ethics approval and consent to participate:
Not applicable.
\item Consent for publication:
Not applicable.
\item Data availability:
All training data was collected from the UIUC airfoil database at \cite{Selig1996}.
\item Materials availability:
Not applicable.
\item Code availability:
Code used in this publication is available at \url{https://github.com/gravesreid/Airfoil_Diffusion_Generator.git}.
\item Author contribution:
\begin{itemize}
    \item \textbf{Reid Graves}: Methodology, Software, Data Curation, Analysis, Writing-- Original Draft, Visualization, Writing-- Review and Editing.
    \item \textbf{Amir Barati Farimani}: Conceptualization, Supervision, Writing-- Review and Editing.
\end{itemize}
\end{itemize}

\bibliography{references}

\begin{thebibliography}{10}
\expandafter\ifx\csname url\endcsname\relax
  \def\url#1{\burl{#1}}\fi
\expandafter\ifx\csname urlprefix\endcsname\relax\def\urlprefix{URL }\fi
\providecommand{\bibinfo}[2]{#2}
\providecommand{\eprint}[2][]{\url{#2}}
\providecommand{\doi}[1]{\url{https://doi.org/#1}}
\bibcommenthead

\bibitem{review_optimization_methods}
\bibinfo{author}{Wang, L.} \emph{et~al.}
\newblock \bibinfo{title}{A review of intelligent airfoil aerodynamic optimization methods based on data-driven advanced models}.
\newblock \emph{\bibinfo{journal}{Mathematics}} \textbf{\bibinfo{volume}{12}} (\bibinfo{year}{2024}).
\newblock \urlprefix\url{https://www.mdpi.com/2227-7390/12/10/1417}.

\bibitem{raymer2006aircraft}
\bibinfo{author}{Raymer, D.~P.}
\newblock \emph{\bibinfo{title}{Aircraft Design: A Conceptual Approach}} \bibinfo{edition}{4} edn.
\newblock AIAA Education Series (\bibinfo{publisher}{American Institute of Aeronautics and Astronautics}, \bibinfo{year}{2006}).

\bibitem{Wang2021Airfoil}
\bibinfo{author}{Wang, Y.}, \bibinfo{author}{Shimada, K.} \& \bibinfo{author}{Barati~Farimani, A.}
\newblock \bibinfo{title}{Airfoil gan: encoding and synthesizing airfoils for aerodynamic shape optimization}.
\newblock \emph{\bibinfo{journal}{J Comput Des Eng}} \textbf{\bibinfo{volume}{10}}, \bibinfo{pages}{1350--1362} (\bibinfo{year}{2023}).
\newblock \urlprefix\url{https://doi.org/10.1093/jcde/qwad046}.

\bibitem{Morris2008Wing}
\bibinfo{author}{Morris, A.}, \bibinfo{author}{Allen, C.} \& \bibinfo{author}{Rendall, T.}
\newblock \emph{\bibinfo{title}{Wing Design by Aerodynamic and Aeroelastic Shape Optimisation}}.
\newblock \urlprefix\url{https://arc.aiaa.org/doi/abs/10.2514/6.2008-7054}.
\newblock \eprint{https://arc.aiaa.org/doi/pdf/10.2514/6.2008-7054}.

\bibitem{Hermans2014Automated}
\bibinfo{author}{Hermans, M.}, \bibinfo{author}{Schrauwen, B.}, \bibinfo{author}{Bienstman, P.} \& \bibinfo{author}{Dambre, J.}
\newblock \bibinfo{title}{Automated design of complex dynamic systems}.
\newblock \emph{\bibinfo{journal}{PLOS ONE}} \textbf{\bibinfo{volume}{9}}, \bibinfo{pages}{1--11} (\bibinfo{year}{2014}).
\newblock \urlprefix\url{https://doi.org/10.1371/journal.pone.0086696}.

\bibitem{jadhav2023stressd}
\bibinfo{author}{Jadhav, Y.} \emph{et~al.}
\newblock \bibinfo{title}{Stressd: 2d stress estimation using denoising diffusion model}.
\newblock \emph{\bibinfo{journal}{Comput Methods Appl Mech Eng}} \textbf{\bibinfo{volume}{416}}, \bibinfo{pages}{116343} (\bibinfo{year}{2023}).

\bibitem{jadhav2024generative}
\bibinfo{author}{Jadhav, Y.} \emph{et~al.}
\newblock \bibinfo{title}{Generative lattice units with 3d diffusion for inverse design: Glu3d}.
\newblock \emph{\bibinfo{journal}{Adv Funct Mater}} \bibinfo{pages}{2404165} (\bibinfo{year}{2024}).

\bibitem{Chen2021Deep}
\bibinfo{author}{Chen, W.} \& \bibinfo{author}{Ramamurthy, A.}
\newblock \emph{\bibinfo{title}{Deep Generative Model for Efficient 3D Airfoil Parameterization and Generation}}.
\newblock \urlprefix\url{https://arc.aiaa.org/doi/abs/10.2514/6.2021-1690}.
\newblock \eprint{https://arc.aiaa.org/doi/pdf/10.2514/6.2021-1690}.

\bibitem{Xie2022Soft}
\bibinfo{author}{Xie, H.}, \bibinfo{author}{Wang, J.} \& \bibinfo{author}{Zhang, M.}
\newblock \bibinfo{title}{Parametric generative schemes with geometric constraints for encoding and synthesizing airfoils}.
\newblock \emph{\bibinfo{journal}{Eng Appl Artif Intell}} \textbf{\bibinfo{volume}{128}}, \bibinfo{pages}{12} (\bibinfo{year}{2024}).
\newblock \urlprefix\url{https://doi.org/10.1016/j.engappai.2023.107505}.

\bibitem{Barrett2006Airfoil}
\bibinfo{author}{Barrett, T.~R.}, \bibinfo{author}{Bressloff, N.} \& \bibinfo{author}{Keane, A.}
\newblock \bibinfo{title}{Airfoil shape design and optimization using multifidelity analysis and embedded inverse design}.
\newblock \emph{\bibinfo{journal}{AIAA J}} \textbf{\bibinfo{volume}{44}}, \bibinfo{pages}{2051--2060} (\bibinfo{year}{2006}).

\bibitem{Zhang2020BezierOptimization}
\bibinfo{author}{Zhang, Y.} \& \bibinfo{author}{Joo, S.}
\newblock \bibinfo{title}{Optimizing airfoil shape using b-spline curves}.
\newblock \emph{\bibinfo{journal}{Comput Fluids}} \textbf{\bibinfo{volume}{201}}, \bibinfo{pages}{104468} (\bibinfo{year}{2020}).

\bibitem{Mukesh2021BSpline}
\bibinfo{author}{Mukesh, A.} \& \bibinfo{author}{Rathakrishnan, E.}
\newblock \bibinfo{title}{Optimization of airfoil shapes using b-spline curves and genetic algorithms}.
\newblock \emph{\bibinfo{journal}{Aerosp Sci Technol}} \textbf{\bibinfo{volume}{110}}, \bibinfo{pages}{106501} (\bibinfo{year}{2021}).

\bibitem{Li2020GeometricGAN}
\bibinfo{author}{Li, J.}, \bibinfo{author}{Zhang, M.} \& \bibinfo{author}{Martins, J. R. R.~A.}
\newblock \bibinfo{title}{Efficient aerodynamic shape optimization with deep-learning-based geometric filtering}.
\newblock \emph{\bibinfo{journal}{AIAA J}} \textbf{\bibinfo{volume}{58}}, \bibinfo{pages}{4243--4259} (\bibinfo{year}{2020}).

\bibitem{Wei2023Diffusion}
\bibinfo{author}{Wei, Z.}, \bibinfo{author}{Guillard, B.}, \bibinfo{author}{Bauerheim, M.}, \bibinfo{author}{Chapin, V.} \& \bibinfo{author}{Fua, P.}
\newblock \bibinfo{title}{Latent representation of cfd meshes and application to 2d airfoil aerodynamics}.
\newblock \emph{\bibinfo{journal}{AIAA J}}  (\bibinfo{year}{2023}).

\bibitem{DiffAirfoil}
\bibinfo{author}{Wei, Z.}, \bibinfo{author}{Dufour, E.}, \bibinfo{author}{Pelletier, C.}, \bibinfo{author}{Fua, P.} \& \bibinfo{author}{Bauerheim, M.}
\newblock \bibinfo{title}{Diffairfoil: An efficient novel airfoil sampler based on latent space diffusion model for aerodynamic shape optimization}.
\newblock \emph{\bibinfo{journal}{AIAA AVIATION Forum}}  (\bibinfo{year}{2024}).
\newblock \urlprefix\url{https://infoscience.epfl.ch/server/api/core/bitstreams/9025926a-858f-422c-82d3-134bdb8cb759/content}.

\bibitem{mangalam2021overcoming}
\bibinfo{author}{Mangalam, K.} \& \bibinfo{author}{Garg, R.}
\newblock \bibinfo{title}{Overcoming mode collapse with adaptive multi adversarial training}  (\bibinfo{year}{2021}).

\bibitem{bayat2023a}
\bibinfo{author}{Bayat, R.}
\newblock \bibinfo{title}{A study on sample diversity in generative models: {GAN}s vs. diffusion models} (\bibinfo{year}{2023}).
\newblock \urlprefix\url{https://openreview.net/forum?id=BQpCuJoMykZ}.

\bibitem{Selig1996}
\bibinfo{author}{Selig, M.}
\newblock \bibinfo{title}{Uiuc airfoil data site} (\bibinfo{year}{1996}).
\newblock \urlprefix\url{http://m-selig.ae.illinois.edu/ads/coord_database.html}.
\newblock \bibinfo{note}{Department of Aeronautical and Astronautical Engineering, University of Illinois at Urbana-Champaign}.

\bibitem{aerosandbox}
\bibinfo{author}{Sharpe, P.~D.}
\newblock \emph{\bibinfo{title}{AeroSandbox: A Differentiable Framework for Aircraft Design Optimization}}.
\newblock Master's thesis, \bibinfo{school}{Massachusetts Institute of Technology} (\bibinfo{year}{2021}).

\bibitem{koning2023improved}
\bibinfo{author}{Koning, W. J.~F.}, \bibinfo{author}{Romander, E.~A.}, \bibinfo{author}{Cummings, H.~V.}, \bibinfo{author}{Perez~Perez, B.~N.} \& \bibinfo{author}{Buning, P.~G.}
\newblock \bibinfo{title}{On improved understanding of airfoil performance evaluation methods at low reynolds numbers}.
\newblock \emph{\bibinfo{journal}{J Aircr}}  (\bibinfo{year}{2023}).

\bibitem{gracey1958summary}
\bibinfo{author}{Gracey, W.}
\newblock \bibinfo{title}{Summary of methods of measuring angle of attack on aircraft}.
\newblock \bibinfo{type}{Technical Note} \bibinfo{number}{4351}, \bibinfo{institution}{National Advisory Committee for Aeronautics} (\bibinfo{year}{1958}).
\newblock \urlprefix\url{https://ntrs.nasa.gov/citations/19930085167}.

\bibitem{neuralfoil}
\bibinfo{author}{Sharpe, P.}
\newblock \bibinfo{title}{Neuralfoil: An airfoil aerodynamics analysis tool using physics-informed machine learning}.
\newblock \bibinfo{howpublished}{\url{https://github.com/peterdsharpe/NeuralFoil}} (\bibinfo{year}{2023}).

\bibitem{Croitoru2022Diffusion}
\bibinfo{author}{Croitoru, F.-A.}, \bibinfo{author}{Hondru, V.}, \bibinfo{author}{Ionescu, R.~T.} \& \bibinfo{author}{Shah, M.}
\newblock \bibinfo{title}{Diffusion models in vision: A survey}.
\newblock \emph{\bibinfo{journal}{IEEE Trans Pattern Anal Mach Intell}} \textbf{\bibinfo{volume}{45}}, \bibinfo{pages}{10850--10869} (\bibinfo{year}{2022}).

\bibitem{Ho2021Cascaded}
\bibinfo{author}{Ho, J.} \emph{et~al.}
\newblock \bibinfo{title}{Cascaded diffusion models for high fidelity image generation}.
\newblock \emph{\bibinfo{journal}{J Mach Learn Res}} \textbf{\bibinfo{volume}{23}}, \bibinfo{pages}{47:1--47:33} (\bibinfo{year}{2021}).

\bibitem{ho2020denoising}
\bibinfo{author}{Ho, J.}, \bibinfo{author}{Jain, A.} \& \bibinfo{author}{Abbeel, P.}
\newblock \bibinfo{editor}{.} (ed.) \emph{\bibinfo{title}{Denoising diffusion probabilistic models}}.
\newblock (ed.\bibinfo{editor}{.}) \emph{\bibinfo{booktitle}{Proceedings of the 34th International Conference on Neural Information Processing Systems (NIPS)}}, NIPS '20 (\bibinfo{publisher}{Curran Associates Inc.}, \bibinfo{address}{Red Hook, NY, USA}, \bibinfo{year}{2020}).

\bibitem{ogoke2023inexpensive}
\bibinfo{author}{Ogoke, F.} \emph{et~al.}
\newblock \bibinfo{title}{Inexpensive high fidelity melt pool models in additive manufacturing using generative deep diffusion}.
\newblock \emph{\bibinfo{journal}{Mater Des}} \textbf{\bibinfo{volume}{245}}, \bibinfo{pages}{113181} (\bibinfo{year}{2024}).
\newblock \urlprefix\url{https://www.sciencedirect.com/science/article/pii/S0264127524005562}.

\bibitem{Yang2022Diffusion}
\bibinfo{author}{Yang, L.} \emph{et~al.}
\newblock \bibinfo{title}{Diffusion models: A comprehensive survey of methods and applications}.
\newblock \emph{\bibinfo{journal}{ACM Comput Surv}} \textbf{\bibinfo{volume}{56}}, \bibinfo{pages}{1--39} (\bibinfo{year}{2022}).

\bibitem{sohldickstein2015deep}
\bibinfo{author}{Sohl-Dickstein, J.}, \bibinfo{author}{Weiss, E.}, \bibinfo{author}{Maheswaranathan, N.} \& \bibinfo{author}{Ganguli, S.}
\newblock \bibinfo{editor}{.} (ed.) \emph{\bibinfo{title}{Deep unsupervised learning using nonequilibrium thermodynamics}}.
\newblock (ed.\bibinfo{editor}{.}) \emph{\bibinfo{booktitle}{Proceedings of the 32nd International Conference on Machine Learning}}, Vol.~\bibinfo{volume}{37} of \emph{\bibinfo{series}{Proceedings of Machine Learning Research}}, \bibinfo{pages}{2256--2265} (\bibinfo{publisher}{PMLR}, \bibinfo{address}{Lille, France}, \bibinfo{year}{2015}).
\newblock \urlprefix\url{https://proceedings.mlr.press/v37/sohl-dickstein15.html}.

\bibitem{degroot2014probability}
\bibinfo{author}{DeGroot, M.~H.} \& \bibinfo{author}{Schervish, M.~J.}
\newblock \emph{\bibinfo{title}{Probability and Statistics}} \bibinfo{edition}{4} edn (\bibinfo{publisher}{Pearson}, \bibinfo{address}{Boston}, \bibinfo{year}{2014}).

\bibitem{ronneberger2015unetconvolutionalnetworksbiomedical}
\bibinfo{author}{Ronneberger, O.}, \bibinfo{author}{Fischer, P.} \& \bibinfo{author}{Brox, T.}
\newblock \bibinfo{editor}{.} (ed.) \emph{\bibinfo{title}{U-net: Convolutional networks for biomedical image segmentation}}.
\newblock (ed.\bibinfo{editor}{.}) \emph{\bibinfo{booktitle}{Medical Image Computing and Computer-Assisted Intervention (MICCAI)}}, Vol. \bibinfo{volume}{9351} of \emph{\bibinfo{series}{LNCS}}, \bibinfo{pages}{234--241} (\bibinfo{publisher}{Springer}, \bibinfo{year}{2015}).
\newblock \urlprefix\url{http://lmb.informatik.uni-freiburg.de/Publications/2015/RFB15a}.
\newblock \bibinfo{note}{Available on arXiv:1505.04597 [cs.CV]}.

\bibitem{ho2022classifierfreediffusionguidance}
\bibinfo{author}{Ho, J.} \& \bibinfo{author}{Salimans, T.}
\newblock \bibinfo{editor}{.} (ed.) \emph{\bibinfo{title}{Classifier-free diffusion guidance}}.
\newblock (ed.\bibinfo{editor}{.}) \emph{\bibinfo{booktitle}{NeurIPS 2021 Workshop on Deep Generative Models and Downstream Applications}} (\bibinfo{year}{2021}).
\newblock \urlprefix\url{https://openreview.net/forum?id=qw8AKxfYbI}.

\bibitem{hu2024drugdiscoverysmilestopharmacokineticsdiffusion}
\bibinfo{author}{Hu, B.}, \bibinfo{author}{Layton, A.} \& \bibinfo{author}{Chen, H.}
\newblock \bibinfo{title}{Drug discovery smiles-to-pharmacokinetics diffusion models with deep molecular understanding} (\bibinfo{year}{2024}).
\newblock \urlprefix\url{https://arxiv.org/abs/2408.07636}.
\newblock \eprint{2408.07636}.

\bibitem{bsplines}
\bibinfo{author}{Dierckx, P.}
\newblock \emph{\bibinfo{title}{Curve and Surface Fitting with Splines}}  (\bibinfo{publisher}{Oxford University Press}, \bibinfo{year}{1993}).
\newblock \urlprefix\url{https://doi.org/10.1093/oso/9780198534419.001.0001}.

\bibitem{DIERCKX1975165}
\bibinfo{author}{Dierckx, P.}
\newblock \bibinfo{title}{An algorithm for smoothing, differentiation and integration of experimental data using spline functions}.
\newblock \emph{\bibinfo{journal}{Journal of Computational and Applied Mathematics}} \textbf{\bibinfo{volume}{1}}, \bibinfo{pages}{165--184} (\bibinfo{year}{1975}).
\newblock \urlprefix\url{https://www.sciencedirect.com/science/article/pii/0771050X75900340}.

\bibitem{2020SciPy-NMeth}
\bibinfo{author}{Virtanen, P.} \emph{et~al.}
\newblock \bibinfo{title}{{{SciPy} 1.0: Fundamental Algorithms for Scientific Computing in Python}}.
\newblock \emph{\bibinfo{journal}{Nature Methods}} \textbf{\bibinfo{volume}{17}}, \bibinfo{pages}{261--272} (\bibinfo{year}{2020}).

\bibitem{yonekuraPCA}
\bibinfo{author}{Yonekura, K.} \& \bibinfo{author}{Watanabe, O.}
\newblock \bibinfo{title}{A shape parameterization method using principal component analysis in applications to parametric shape optimization}.
\newblock \emph{\bibinfo{journal}{Journal of Mechanical Design}} \textbf{\bibinfo{volume}{136}}, \bibinfo{pages}{121401} (\bibinfo{year}{2014}).
\newblock \urlprefix\url{https://doi.org/10.1115/1.4028273}.

\bibitem{mackiewicz1993principal}
\bibinfo{author}{Ma{\'c}kiewicz, A.} \& \bibinfo{author}{Ratajczak, W.}
\newblock \bibinfo{title}{Principal components analysis (pca)}.
\newblock \emph{\bibinfo{journal}{Computers \& Geosciences}} \textbf{\bibinfo{volume}{19}}, \bibinfo{pages}{303--342} (\bibinfo{year}{1993}).

\bibitem{akram2021aerodynamic}
\bibinfo{author}{Akram, M.~T.} \& \bibinfo{author}{Kim, M.-H.}
\newblock \bibinfo{title}{Aerodynamic shape optimization of nrel s809 airfoil for wind turbine blades using reynolds-averaged navier stokes model—part ii}.
\newblock \emph{\bibinfo{journal}{Appl Sci}} \textbf{\bibinfo{volume}{11}}, \bibinfo{pages}{2211} (\bibinfo{year}{2021}).

\end{thebibliography}

\end{document}